\documentclass{article}
\usepackage[utf8]{inputenc}
\usepackage{hyperref}

\title{Annotated Dataset Creation through General Purpose Language Models for non-English Medical NLP}
\author{
  Johann Frei
  \and
  Frank Kramer
  \and
  \texttt{firstname.lastname@informatik.uni-augsburg.de}
}
\date{August 2022}

\usepackage{xcolor}
\usepackage{listings}
\usepackage{graphicx}
\usepackage{multicol}
\usepackage{multirow}

\begin{document}

\maketitle
\begin{abstract}
Obtaining text datasets with semantic annotations is an effortful process, yet crucial for supervised training in natural language processsing (NLP). In general, developing and applying new NLP pipelines in domain-specific contexts for tasks often requires custom designed datasets to address NLP tasks in supervised machine learning fashion. When operating in non-English languages for medical data processing, this exposes several minor and major, interconnected problems such as lack of task-matching datasets as well as task-specific pre-trained models.

In our work we suggest to leverage pretrained language models for training data acquisition in order to retrieve sufficiently large datasets for training smaller and more efficient models for use-case specific tasks. To demonstrate the effectiveness of your approach, we create a custom dataset which we use to train a medical NER model for German texts, GPTNERMED, yet our method remains language-independent in principle. Our obtained dataset as well as our pre-trained models are publicly available at \url{https://github.com/frankkramer-lab/GPTNERMED}.
\end{abstract}

\section{Introduction}
In situations of low resource languages, neural baseline techniques for specific tasks in natural language processing (NLP) are often difficult to be applied successfully due to the lack of sufficient and adequately annotated training data. While English can be perceived as the most relevant language in the field of NLP research as being a high-resource language, effectively any other language can be considered as rather low-resource language in contrast. Yet, the abundance of plain textual resources is no uniquely decisive factor when it comes to dealing with embedded NLP problems in real-life applications. In this regard, a domain-specific dataset needs to be obtained to match to the applied context and the underlying data acquisition process can involve access to highly restricted data, manual engagements from domain experts or time- and cost-intensive data gathering. Another concern relates to the actual NLP objective of the use case and usually heavily determines the final design of the obtained dataset and its collection of task-related annotations.

We study the use case to annotate certain medical entity classes in German throughout this paper since it is an instance that suffers from all formerly mentioned challenges. In this work, we demonstrate an effective method for synthesizing a custom, domain-aligned dataset with annotation information in an unsupervised fashion. Furthermore, we show evidence of its effectiveness by training a generic medical model for German medical named entity recognition (NER) by finetuning a pre-trained language model. Due to the inherently generic nature of our work, we do not see fundamental obstacles in apply the approach on related entity classes in medical or even non-medical tasks, or for different non-English languages of similar quantitative level of resource abundance.

\section{Background and Related Work}
\subsection{Medical Datasets}
In NLP, deep learning-based methods have been proven as highly effective in order to tackle frequent tasks, most notably the self-attention-mechanism-based transformer architecture.\cite{vaswani_attention_2017} One fundamental problem of deep learning-based methods remains to be the need for vast amount of data for training, including corresponding annotations for supervised learning.

In English medical NLP, these challenges have been addressed to a certain extend by the availability of annotated datasets, such as the MIMIC-III\cite{pollard_mimic-iii_2016} and MIMIC-IV\cite{johnson_mimic-iv_2021} datasets or n2c2 datasets from the i2b2 challenges\cite{henry_2018_2020}.
In general, multilingual textual datasets are available that carry medical texts from multiple languages. The datasets often entail parallel corpora for translation tasks and lack semantic annotation like the \textit{UFAL Medical Corpus}\footnote{UFAL Medical Corpus (acceessed at 22.08.2022): \url{https://ufal.mff.cuni.cz/ufal_medical_corpus}} for the WMT'17 biomedical challenge\cite{yepes_findings_2017}. Driven by manual annotation work, Mantra GSC\cite{kors_multilingual_2015} is a public gold-standard annotated corpus with multilingual texts based on prior parallel corpora and provides limited UMLS information.

For German medical NLP, the field has made notable advances in terms of available datasets. While work in this field of NLP has been published, internal and proprietary datasets are frequently used as underlying datasets.\cite{hahn_3000pa-towards_2018,roller_fine-grained_2016,wermter_annotated_2004,kreuzthaler_detection_2015,toepfer_fine-grained_2015,bretschneider_identifying_2013,fette_information_2012,konig_knowledge-based_2019,cotik_negation_2016,lohr_operative_1992,minarro-gimenez_quantitative_2019,krebs_semi-automatic_2017} In recent years, semi-publicly available datasets like BRONCO\cite{kittner_annotation_2021} and GGPONC 1.0\cite{borchert_ggponc_2022} and 2.0\cite{borchert_ggponc_2022} have been made available. While BRONC is advertised based on real discharge letters with annotations, other datasets like GGPONC originate from synthetic data sources like clinical practice guidelines, assembled from multiple or crawled data from the web. If annotation data is provided, such metadata differ in terms of entity types, entity type definitions or their overall task objectives. Hence, a direct comparison of datasets and corresponding models cannot be made directly with respect to NER F1/tagging scores, or entity linking to different onthologies. Only metrics of rather limited interest such as test set performance of trained models, or token size and number of entities for a dataset are directly derivable for comparison.
For an extensive overview on the recent state of German medical datasets, we point to \cite{borchert_ggponc_2022}.

\subsection{Medical Models and Applications}
We restrict our focus on models and applications to items of general interest and practical applicability.
Most works from the presented dataset section develop accompanied models to the datasets and publish internally evaluated scores. However in many cases, the reproducibility of the described results is not possible since models are not made publicly available along with the paper. Furthermore, some models or systems are designed for narrow NLP tasks and are not of interest for general application in the field, like cardiography texts\cite{toepfer_fine-grained_2015}. Since models are trained on sensitive training data, privacy concerns arise from the fact that potential training data extraction attacks could uncover patient-related data. This concern is amplified by the increasing use of fine-tuning larger language models that are susceptible to such attacks\cite{carlini_extracting_2021}. In the German domain, the neural German model GERNERMED\cite{frei_gernermed_2022} avoids this issue by using public data from English in combination with neural machine translation to be the first publicly available model with unrestricted access and further improved their method for stronger models\cite{frei_gernermed_2022-1}. Authors from GGPONC\cite{borchert_ggponc_2022} and BRONCO\cite{kittner_annotation_2021} provide access to their own models after registration or signed user agreement. On a broader perspective, the software mEx\cite{roller_mex_2018} provides a entire stack of different models and dockerized software layers to serve an integrated text processing system, their models can be obtained on request through signed user agreement\cite{roller_medical_2022}. Commercial applications from Health Discovery (Averbis)\footnote{\url{https://averbis.com/de/health-discovery/}} and SparkNLP (John Snow labs)\footnote{\url{https://nlp.johnsnowlabs.com/analyze_medical_text_german}} are available but are purely proprietary applications. Contrary to perceptions of domain experts and reviewers, Amazon Comprehend Medical\footnote{\url{https://docs.aws.amazon.com/comprehend-medical/latest/dev/comprehendmedical-welcome.html}} does \textit{not} support German texts at the time of writing. Popular, open solutions like Apache cTAKES\cite{savova_mayo_2010} and MetaMaps\cite{aronson_overview_2010} do not exist for the German community. Due to the rapid change in the field, we do not consider this list of available models and software as conclusive. We point to \cite{roller_medical_2022} and \cite{borchert_ggponc_2022} for a more exhaustive enumeration of available models and systems.

\subsection{Language Model-based Dataset Generation}
Data augmentation is a popular technique in the Machine Learning community, in which the objective is to sample new data points from the manifold that models the set of known data points. In computer vision, semantic invariance applies to basic image transformation in many situations\cite{simard_transformation_1998}. However in NLP such basic techniques cannot always be applied if semantic information of sentences needs to be preserved, but more sophisticated approaches are used such as back-translation\cite{sennrich_improving_2016} of words or phrases through translation, yet failures in translation can jeopardize the augmentation method\cite{artetxe_translation_2020}. The idea to use pretrained language models for data augmentation has been proposed as effective method for augmenting small datasets\cite{anaby-tavor_not_2020, raille_fast_2020} or even create datasets nearly from scratch.\cite{schick_few-shot_2021, schick_generating_2021}. With the increasing popularity of large, prompt-based language models like GPT-2/3\cite{radford_language_2019,brown_language_2020} and open source counterparts\cite{wang_gpt-j-6b_2021, black_gpt-neox-20b_2022}, methods with various objectives have been developed to improve the quality and usefulness of the models in different contexts such as sentence similarity estimation\cite{schick_generating_2021}. In addition to classical few-shot text generation, task instruction-driven zero-shot methods are likewise an active field of research\cite{puri_zero-shot_2019, meng_generating_2022, schick_generating_2021}.
For medical NLP purposes, text generation has been shown for synthesizing EHR reports\cite{libbi_generating_2021} and its application for downstream tasks\cite{amin-nejad_exploring_2020} using an GPT-2 model. To the best of our knowledge, we are the first team to expand the general idea to the field of German medical NLP.

\section{Methods}

In this work, we leverage the capabilities of pre-trained language models in regards to their example-driven few-shot learning for text generation. The method follows the basic idea implemented in various related contexts\cite{libbi_generating_2021,anaby-tavor_not_2020,schick_generating_2021}. We apply the GPT NeoX language model from EleutherAI\cite{black_gpt-neox-20b_2022} for input processing and text generation. The model implementation is kept close to the GPT-2/3 architecture, an autoregressive model which is closely related to the vanilla Transformer architecture\cite{vaswani_attention_2017} with decoder-only blocks. Note that we do not perform gradient-based fine-tuning of the model on novel data, but the model is only used for inference. In difference to other models like GPT-3 (\cite{brown_language_2020}), the internal model weights are publicly available similar to its smaller GPT-J\cite{wang_gpt-j-6b_2021} model. We decided to use the NeoX model over GPT-J due to its larger size\footnote{Model parameter size (billions): GPT-J: 6B, GPT NeoX: 20B, GPT-3: 175B} which has been shown to exceed the performance of GPT-J on several tasks\cite{black_gpt-neox-20b_2022} yet being sufficiently small to run on our local instance. In addition, large multilingual language models are able to improve task performance on low-resource languages (e.g. German) by the multilingual knowledge transfer from a high-resource language (e.g. English)\cite{schuster_cross-lingual_2019}.

As previously discussed, LM-based text generation models are used to generate their respective text output by conditioning on an input text sequence, highlighting two main aspects on the input sequence design. First, the sequence can carry a task description in natural language to advise the model on its task objective. While writing an obvious prompt command seems obvious to a normal person, the performance of language models vary between different semantically equivalent task instructions\cite{schick_generating_2021}. Second, the input can inject information on the task during the prediction of the next word by providing text examples in the input sequence.

\begin{figure}
    \centering
    \includegraphics[width=0.6\textwidth]{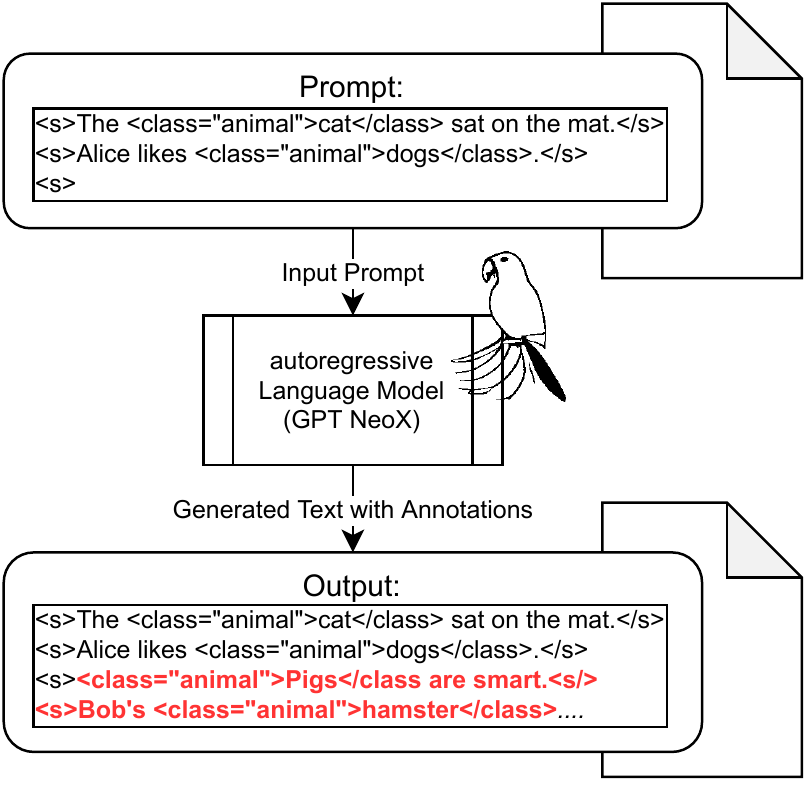}
    \caption{Synthesis of markup-based text with annotation information: The input prompt consists of markup-encoded text of few pre-written sentences. The set of sentences is augmented by the language model that generates new data (red) token-wise in an autoregressive fashion.}
    \label{fig:markup_simple}
\end{figure}

In this work, we do not focus on tuning task instructions in natural language but rather demonstrate that straightforward few-shot-learning-based example prompting as model input suffices for synthetic dataset generation within the scope of our use case. To avoid the issue of only generating plain natural text without valuable annotation metadata, we design our input prompt in the style of a simple markup language, where the language model reads the data as a collection of sentences. Each sentence is enclosed by \texttt{<s>} and \texttt{</s>} signs and separated by a line brake. For each sentence every word from a certain label class $l$ is enclosed by \texttt{<class="$l$">} and \texttt{</class>} respectively. We select a small set of exemplary sentences, encode them according to the basic markup rules and append the opening sentence tag \texttt{<s>} to the prompt to indicate the start of an additional sentence. The whole process is illustrated in Figure \ref{fig:markup_simple}.

In language generation, the unnormalized probabilities over tokens, referred to as logits, are normalized and smoothed by the last softmax layer in the network

\begin{equation}
    softmax(l_i) = \frac{e^{l_i/\tau}}{\sum^{n}_{j}e^{l_j/\tau}}
\end{equation}

where $n$ is the number of tokens in the vocabulary, $l_i$ is the unnormalized predicted probability for token $i$. The temperature parameter $\tau$ is used for smoothing the normalized probability distribution. In this, higher values of $\tau$ increase the probabilities for less probable tokens at the expense of highly probable tokens. We can utilize the parameter to reduce the risk of generating invalid markup-based text data by setting the temperature to $0 < \tau < 1.0$ in combination with $\textit{top-p} < 1.0$ prior to token sampling.

After collecting the output data, we parse the markup text to obtain a synthetic, silver-standard corpus along with its corresponding annotations. For a further data cleansing, we only keep sentences that fulfill the following requirements: First, the sentence needs to have a closing \texttt{</s>} tag. Second, the parsing of the sentence can succeed and the annotations are provided by valid \texttt{<class="$l$">} and \texttt{</class>} tags. Third, the sentence has at least one annotation. Fourth, all annotation labels are part of the pre-defined set of label classes. Fifth, duplicate sentences are reduced to unique occurrences (deduplication).

The synthesis of annotated sentences from a large language model and its transfer to a smaller, more efficient model can be considered as a high level form of knowledge distillation: For the very purpose of developing a German NER model for medical entities, we are able to transform the implicit knowledge of the 20B parameter model about this very context into a dedicated NER model with an faster, less resource-intensive computational footprint. In fact, these properties align well to the aim of practical applicability of our method and its resulting model in dedicated domain contexts.
For the development of a robust NER model, we train a neural-based NER parser from the open-source SpaCy NLP library on our dataset. While the NER parser component is trained from scratch, its input vectors are generated through an pre-trained BERT-based encoder model to improve the performance of the final model through transfer-learning and contextualization. The BERT-based encoder is fine-tuned to the data by gradient update during training procedure.

\section{Results}

We provide the GPT-based NeoX model with an input sequence of twelve sentences in German language, encoded in the described markup style. The sentences are pre-annotated with the label classes \textit{Medikation} (medication/drug), \textit{Dosis}(dosage/strength) and \textit{Diagnose} (diagnosis). The prompt is displayed in Figure \ref{fig:prompt}.

\begin{figure}[!ht]
    {
    \UseRawInputEncoding
    \begin{lstlisting}[language={}, breaklines=true, frame=single, basicstyle=\scriptsize\ttfamily, keepspaces=true,numbers=left, columns=flexible, postbreak=\mbox{\textcolor{red}{$\hookrightarrow$}\space}]
<s>Zur weiteren Bekämpfung des <class="Diagnose">Juckreiz</class> wird die Einnahme von täglich <class="Dosis">100mg</class> <class="Medikation">Cortison</class> empfohlen.</s>
<s>Bei wiederkehrender Infektion wie einer <class="Diagnose">Sepsis</class> oder schweren <class="Diagnose">Pnseumonien</class> wird eine Überwachung erforderlich sein.</s>
<s><class="Medikation">Valsartan</class>/<class="Medikation">HCT</class> <class="Dosis">160</class>/<class="Dosis">12,5 mg</class> 1-0-0</s>
<s><class="Medikation">Pantoprazol</class> <class="Dosis">40 mg</class> p.o.</s>
<s>Die feingewebliche histopathologische Untersuchung ergab den Befund einer <class="Diagnose">Metastase</class> des bekannten malignen <class="Diagnose">Melanoms</class>.</s>
<s><class="Diagnose">Diabetes Typ 2</class>-Patienten müssen regelmäßig <class="Medikation">Insulin</class> (mindestens mit <class="Dosis">12ml</class> dosiert) spritzen.</s>
<s>Ich nehme <class="Medikation">Antibiotika</class> seit Tagen. Seitdem ist die <class="Diagnose">Mandelentzündung</class> deutlich besser geworden.</s>
<s>Entlassung: <class="Dosis">40mg</class> <class="Medikation">Lidocain</class> wegen <class="Diagnose">Kopfschmerzen</class></s>
<s>Zusammenfassende D: Zervix-PE bei 11 und 2 Uhr mit ausgeprägter <class="Diagnose">chronisch-florider Zervizitis<class="Diagnose">.</s>
<s>Die Verschreibung von <class="Medikation">Hämatokrin</class> <class="Dosis">43mg</class> war unnötig.</s>
<s>Der Patient klagt über <class="Diagnose">Karditiden</class> und nimmt täglich <class="Medikation">Nifedipin</class> ein.</s>
<s>D: PE-Material der Portio bei 1 Uhr mit Nachweis einer schwergradigen <class="Diagnose">squamösen intraepithelialen Läsion</class> (<class="Diagnose">HSIL</class>; hier noch <class="Diagnose">CIN II</class>).</s>
<s>\end{lstlisting}
    }
    \caption{Input prompt: The sentences are encoded according to the markup scheme. The trailing \texttt{<s>} indicates the beginning of a new sentence to the model.}
    \label{fig:prompt}
\end{figure}

During inference we set $\tau$ to 0.8 and $\textit{top-p}$ to 0.9 for language generation and sample 1000 different outputs with a maximum length of 768 tokens each, and additional 100 outputs with an increased temperature $\tau$ set to 0.9. Given the parameters, we obtain a raw \textit{baseline} dataset of 17776 sentences which we reduce to 9845 sentences after the different filters were applied, as shown in Table \ref{tab:dataset_filters}. The final dataset consists of 245107 tokens with annotations for \textit{Dosis} (\# 7547), \textit{Medikation} (\# 9868) and \textit{Diagnose} (\# 5996).

\begin{table}[!ht]
    \centering
    \begin{tabular}{l | r | r | r} 
        \hline
        \textbf{Applied Filter} & \textbf{\#Sentences} & \textbf{\% of Baseline} & \textbf{Impact} \\
        \hline\hline
        Baseline & 17776 & 100\% & \\
        \hline
        $\hookrightarrow$ no \texttt{</s>} tag & 16603 & 93\% & 15\% \\
        $\hookrightarrow$ duplicates removal & 11328 & 64\% & 66\% \\
        $\hookrightarrow$ invalid syntax removal & 11326 & 64\% & 0\% \\
        $\hookrightarrow$ invalid or no labels & 9845 & 55\% & 18\% \\
        \hline
        $\Longrightarrow$ Final & 9845 & 55\% & \\
    \end{tabular}
    \caption{Iterative data cleansing: About half of the predicted sentences have been removed. The majority of sentence removals are due to the duplicate removal filter. All percentage numbers are rounded.}
    \label{tab:dataset_filters}
\end{table}
The inference was computed on an NVIDIA DGX workstation with two NeoX models running in parallel on different A100 GPUs. The inference took a total of 118h of compute, which results in an estimated GPU power consumption of 35,400 Wh and about 15kg of carbon emissions\footnote{According to the United States Environmental Protection Agency: \url{https://www.epa.gov/energy/greenhouse-gas-equivalencies-calculator}}.

As a follow-up step, we train three NER models on the synthesized dataset with pretrained \textbf{gbert-large}\cite{chan_germans_2020}, \textbf{GottBERT-base}\cite{scheible_gottbert_2020} and \textbf{German-MedBERT}\footnote{German MedBERT on Huggingface (accessed 22.08.2022): \url{https://huggingface.co/smanjil/German-MedBERT}} model retrieved from the HuggingFace platform as contextualized feature encoders. We split the dataset randomly into (80\%,10\%,10\%) sets for training, validation and test. The Adam optimizer with an initial learning rate $5e^{-5}$ and a batch size of 128 is used, as we stick close to the default hyperparameters from SpaCy for training. We select the final model based on the lowest F1 score on the validation set. The training iterations took 55m (gbert), 25m (GottBERT), 48m (German-MedBERT).

We evaluate the performance of the respective models on precision, recall and f1-score on the testset. The evaluation is computed in strict mode as a character-wise classification task, meaning that exact overlaps and label classes are considered. The results are shown in Table \ref{tab:results_testset}. The results indicate strong performance of the models on all label classes, with gbert and GottBert as the models with the best averaged f1 scores. As a significant caveat, while the dataset is split into training, validation and test set and no samples are shared across these sets, the synthesized dataset contains structurally similar sentences that allows the models to potentially overfit implicitly by learning syntax and structure of such homogeneous sentences instead of overfit to certain words directly. The homogeneity could be reduced by various techniques including increasing the temperature $\tau$ at the expense of increasing the probability of generating invalid sentences.

\begin{table}[!ht]
    \centering
    \begin{tabular}{lr|rrr|r}
        \hline
        \multicolumn{2}{l|}{\textit{\textbf{Scores on test set}}} & \multicolumn{3}{|c|}{\textbf{NER Tags}} & \textbf{} \\
        \hline
        \textbf{Model} & \textbf{} & \textbf{Medikation} & \textbf{Diagnose} & \textbf{Dosis} & \textbf{Total} \\
        \hline
        \hline
        \multirow{3}{*}{\shortstack[l]{\textbf{gbert-large}}}
        & Pr & 0.870 & 0.870 & 0.883 & 0.918 \\
        & Re & \textbf{0.936} & \textbf{0.895} & \textbf{0.921} & \textbf{0.919} \\
        & F1 & \textbf{0.949} & \textbf{0.882} & \textbf{0.901} & \textbf{0.918} \\
        \hline
        \multirow{3}{*}{\shortstack[l]{\textbf{GottBERT-base}}}
        & Pr & 0.979 & 0.896 & \textbf{0.887} & \textbf{0.936} \\
        & Re & 0.910 & 0.844 & 0.907 & 0.886 \\
        & F1 & 0.943 & 0.870 & 0.897 & 0.910 \\
        \hline
        \multirow{3}{*}{\shortstack[l]{\textbf{German-MedBERT}}}
        & Pr & \textbf{0.980} & \textbf{0.910} & 0.829 & 0.932 \\
        & Re & 0.905 & 0.730 & 0.890 & 0.842 \\
        & F1 & 0.941 & 0.810 & 0.858 & 0.883 \\
        \hline
    \end{tabular}
    \caption{Results on the test set: The total results are based on the labels' frequency-weighted average. The label annotations are evaluated character-wise.}
    \label{tab:results_testset}
\end{table}

We further evaluate the models on a small gold-standard German dataset proposed in \cite{frei_gernermed_2022-1} as an out-of-distribution (OoD) dataset. Since the dataset contains label annotations largely compatible to the n2c2 2018 ADE dataset\cite{henry_2018_2020}, we cannot directly compare all label classes, yet in the interest of an OoD performance evaluation, we assume that the label class \textit{Drug} shares significant semantic overlap with the label class \textit{Medikation}. The results are provided in Table \ref{tab:results_OoD}. Beyond the expected drop in terms of the \textit{Medikation} scores across all models, gbert and GottBERT are identified as the models with the best f1 scores, with GottBERT surpassing gbert by 2.6\% in f1 score (test set reference: -0,6\%).

\begin{table}[!ht]
    \centering
    \begin{tabular}{ll|r}
        \hline
        \multicolumn{2}{l|}{\textit{\textbf{Scores on OoD set}}} & \multicolumn{1}{|c}{\textbf{NER Tag}} \\
        \hline
        \textbf{Model} & \textbf{} & \textbf{Drug $=$ Medikation} \\
        \hline
        \hline
        \multirow{3}{*}{\shortstack[l]{\textbf{gbert-large}}}
        & Pr & 0.707 \\
        & Re & \textbf{0.979} \\
        & F1 & 0.821 \\
        \hline
        \multirow{3}{*}{\shortstack[l]{\textbf{GottBERT-base}}}
        & Pr & \textbf{0.800} \\
        & Re & 0.899 \\
        & F1 & \textbf{0.847} \\
        \hline
        \multirow{3}{*}{\shortstack[l]{\textbf{German-MedBERT}}}
        & Pr & 0.727 \\
        & Re & 0.818 \\
        & F1 & 0.770 \\
        \hline
    \end{tabular}
    \caption{Results on the out-of-distribution dataset: As caveat, the label definitions of \textit{Medikation} (ours) and \textit{Drug}(from the 2018 n2c2 ADE dataset\cite{henry_2018_2020}) is inaccurately assumed to be equivalent for comparison. The label annotations are evaluated character-wise.}
    \label{tab:results_OoD}
\end{table}

\section{Discussion}
We demonstrate the effectiveness of our method for utilizing pre-trained language models for dataset synthesis by training a neural NER model on this dataset, yet the limited availability of annotated German medical NLP datasets with ill-defined or even dissimilar label classes remains a major obstacle when it comes to a more exhaustive, yet reliable evaluation of the trained NER model for all label classes. Given the evaluation scores on the \textit{Drug}/\textit{Medikation} labels it must be considered that our method achieves these results based on twelve initial sentences. Aside from the evaluation, we did not further perform hyperparameter search for dataset synthesis on parameters like temperature $\tau$ or top-k/top-p sampling or beam search due to the high computational costs of running the NeoX model as well as due to limited access to GPU resources. Even though the initial need for computational resources is a major downside of our method, we believe that this factor becomes negligible with respect to the fact that the method can operate without input from costly human annotators. For very domain-specific contexts, such as German medical texts, this not only provides an opportunity to work on NLP approaches independent of external monopoly-like data sources and medical institution that also constitute a severe asymmetry in academic competition. Yet it also allows the further use of the dataset without additional efforts in pseudonymization and legal ramifications that are usually unavoidable when working with datasets originating from real patient data. Therefore, we are able to provide the synthesized corpus and the trained models to third party use publicly without further access restrictions.
While our NER model exhibits strong performance in general and proves the dataset to comprise useful and valid data for text and corresponding annotation, the dataset remains synthetic in nature and thus cannot be considered as an gold standard-level dataset. The question to which degree the corpus carries additional domain knowledge remains open for future work.

\section{Conclusion}
In this work, we leveraged the few-shot ability of the pre-trained language model GPT NeoX to generate an annotated dataset for German medical texts without the need of manual annotations by introducing few annotated text samples to the language model in a simple markup format. We further used the dataset to train NER models by fine-tuning three pre-trained BERT encoder models. Our evaluation on testset as well as OoD set indicates a robust performance of the NER models even for shifts in the dataset. We discussed the disadvantages and advantages of our method as well as its potential implications for the German medical NLP research community and beyond.

The corpus and the trained models are publicly available on GitHub at: \url{https://github.com/frankkramer-lab/GPTNERMED}

\end{document}